\documentclass{article}

\usepackage[final, nonatbib]{neurips_2019}

\usepackage[utf8]{inputenc} 
\usepackage[T1]{fontenc}    
\usepackage{hyperref}       
\usepackage{url}            
\usepackage{booktabs}       
\usepackage{amsfonts}       
\usepackage{nicefrac}       
\usepackage{microtype}      
\usepackage{amsmath}
\usepackage{amssymb}

\title{On the Acceleration of Deep Neural Network Inference using Quantized Compressed Sensing}

\author{%
 Meshia C\'edric Oveneke \thanks{This work is funded by Fit-For-Purpose Technologies. \url{www.fitforpurpose.tech}} \\
  Artificial Intelligence Research Lab\\
  Fit-For-Purpose Technologies \\
  \texttt{cedric.oveneke@fitforpurpose.tech} \\
}

\begin{document}

\maketitle

\begin{abstract}
Accelerating deep neural network (DNN) inference on resource-limited devices is one of the most important barriers to ensuring a wider and more inclusive adoption.
To alleviate this, DNN binary quantization for faster convolution and memory savings is one of the most promising strategies despite its serious drop in accuracy.
The present paper therefore proposes a novel binary quantization function based on quantized compressed sensing (QCS).
Theoretical arguments conjecture that our proposal preserves the practical benefits of standard methods, while reducing the quantization error and the resulting drop in accuracy.
\end{abstract}

\section{Deep Neural Network Inference Acceleration}
Generally there are three methods for accelerating \textit{deep neural network} (DNN) inference: (1) hardware optimization, (2) compiler optimization, and (3) model optimization.
Each of these methods aims at improving the trade-off between computational cost and model accuracy.
In this paper we are interested in model optimization, i.e. architecture optimization \cite{alioscha2020neural,zoph2018learning,tan2019efficientnet}, pruning and compression \cite{chollet2017xception,szegedy2015going,jaderberg2014speeding}, or low-precision arithmetic and quantization \cite{gupta2015deep,jacob2018quantization,pouransari2020least,wang2019haq,qin2020binary,rastegari2016xnor}.
Our particular focus is quantization. 
DNN quantization reduces the size and computational footprint of the models by representing the activations and the weights using a reduced amount of bits. 
We investigate the particular case of binarization, in which the weights are quantized to a single bit.

Given a DNN layer defined as a triplet $\{\mathbf{I}, \mathbf{W},*\}$ representing input tensor, weights tensor and convolution operator respectively, we approximate the convolution using a binary quantization function $Q_B(\cdot)$ as follows:
\begin{equation} \label{eq_BinaryConvolution}
\mathbf{I} * \mathbf{W} \approx \left( Q_B(\mathbf{I}) \oplus Q_B(\mathbf{W})\right) \alpha
\end{equation}
where $*$ denotes the real-valued convolution, $\oplus$ its binary counterpart and $\alpha \in \mathbb{R}^+$ is a trainable scaling factor. 
The main advantage of such an approximation is that the binary convolution can be implemented using bit-wise operations.
This has been reported to result in $58\times$ faster convolution operations and $32\times$ memory savings \cite{rastegari2016xnor}, offering the possibility to run state-of-the-art DNNs on resource-limited devices.
Without loss of generality, we represent the flattened weights of a DNN layer as a vector $\mathbf{w} \in \mathbb{R}^p$ and define the binary weight quantization function as follows:
\begin{equation}\label{eq_DNNQuantization}
Q_B(\mathbf{w}) \triangleq \text{sign}(\mathbf{w})
\end{equation}
resulting into a binary vector $\mathbf{v} = Q_B(\mathbf{w}) \in \{-1,+1\}^p$, with $\mathbf{v}_i = +1$ if $\mathbf{w}_i \geq 0$ and $\mathbf{v}_i = -1$ otherwise, for $i \in [1,p]$.
The function $Q_B$ defined in (\ref{eq_DNNQuantization}) inevitably introduces a significant quantization error and therefore significantly reduces the model accuracy.
To alleviate this, several strategies have been proposed in the literature for finding the optimal scaling factor $\alpha$ in order to compensate the resulting approximation error induced in equation (\ref{eq_BinaryConvolution}).
In \cite{rastegari2016xnor}, the authors proposed the average of the absolute weight values as $\alpha$ value.
Similarly, the work presented in \cite{pouransari2020least} showed that the optimal (in least-squares sense) scaling factor $\alpha$ equals the unbiased estimator of the expectation $\mathbb{E}\left[|\mathbf{w}_i|\right]$.

\section{Quantized Compressed Sensing}
Despite the above-mentioned strategies for quantizing DNNs by finding the optimal scaling factor $\alpha$, none of these strategies is proven to inherently preserve the properties of the original weights.
We therefore advocate for a more principled quantization strategy, based on the well-established \textit{compressed sensing} (CS) techniques \cite{donoho2006compressed}.
CS is about efficiently acquiring and reconstructing signals, with as fundamental premise that certain classes of signals, such as natural images, have a representation in terms of a sparsity inducing basis where most of the coefficients are zero or small and only a few are large \cite{boufounos20081}.
More formally, CS describes how one can recover a signal $\mathbf{w} \in \mathbb{R}^p$ from $m \leq p$ measurements achieved from a sensing matrix $\mathbf{\Phi} \in \mathbb{R}^{m \times p}$ via an under-determined linear system $\mathbf{v} = \mathbf{\Phi}\mathbf{w}$.
For vectors $\mathbf{w}$ restricted to a low-complexity signal set $\mathcal{K} \subset \mathbb{R}^p$, e.g. the set of $k$-sparse vectors, the reconstruction of the signal is guaranteed if $\frac{1}{\sqrt{m}}\mathbf{\Phi}$ respects the \textit{restricted isometry property} (RIP) \cite{xu2020quantized}. 
When $m$ is large enough, the RIP, as introduced in \cite{candes2006robust}, has been proven to be respected with high probability by random matrices \textit{Gaussian random matrices} (GRM)s with entries identically and independently distributed (i.i.d.) as a standard normal distribution $\mathcal{N}(0,1)$ \cite{baraniuk2008simple}.
In the context of DNN weights, we will further assume that $m = p$.

The main contribution of this work is to propose a new binary quantization function $Q^{\text{CS}}_B(\cdot)$, based on the \textit{quantized compressed sensing} (QCS) scheme introduced in \cite{xu2020quantized}.
Similar as for standard CS, when the sensing matrix $\mathbf{\Phi}$ respects the RIP and a pre-quantization bias $\boldsymbol{\xi} \in \mathbb{R}^p$ is added, the following binary quantization function is proven to efficiently acquire and reconstruct the weights $\mathbf{w}$:
\begin{equation}\label{eq_CSQuantization}
Q^{\text{CS}}_B(\mathbf{w}) \triangleq \text{sign}(\mathbf{\Phi}\mathbf{w} + \boldsymbol{\xi})
\end{equation}
where the binary quantization function $\text{sign}(\cdot)$ is applied element-wise to the vector $\mathbf{\Phi}\mathbf{w} + \boldsymbol{\xi}$.
It is further shown in \cite{xu2020quantized} that if we set the pre-quantization bias as a uniform random vector, i.e. $\boldsymbol{\xi} \sim \mathcal{U}^p([0,1])$, it attenuates the impact of $Q^{\text{CS}}_B$ over the linear measurement of $\mathbf{w}$.
Note that the linear measurement of $\mathbf{w}$ only adds a small computational cost of the order $\mathcal{O}(p)$ \cite{liberty2008dense}.
This quantization scheme is also applicable the input tensor $\mathbf{I}$.
As such, we propose an enhanced approximation of the convolution operation defined in equation (\ref{eq_BinaryConvolution}):
\begin{equation} \label{eq_BinaryConvolutionCS}
\mathbf{I} * \mathbf{W} \approx Q^{\text{CS}}_B(\mathbf{I}) \oplus Q^{\text{CS}}_B(\mathbf{W})
\end{equation}
This enhanced approximation guarantees an arbitrary small quantization error, while preserving the important property of resulting into faster convolution operations and drastic memory savings.
Moreover, this new quantization scheme eliminates the need for finding an optimal scaling factor as in (\ref{eq_BinaryConvolution}).
It is also interesting to observe that the standard quantization scheme (\ref{eq_DNNQuantization}) is a particular case of our proposed scheme (\ref{eq_CSQuantization}) when considering the sensing matrix $\mathbf{\Phi}$ to be the identity matrix $\mathbf{I}_p$ and the pre-quantization bias to be $\mathbf{0}$, which doesn't respect the RIP and increases the quantization error.

\section{Conclusion and Future Work}
To the best of our knowledge, this work is one of the first to investigate the synergistic combination of \textit{deep neural network} (DNN) quantization and \textit{quantized compressed sensing} (QCS) for inference acceleration.
By leveraging both paradigms, we've managed to introduce a novel binary quantization function for approximating the convolution.
As a result, our proposal preserves the proven practical benefits in terms of speed of convolution operations and memory savings, while limiting the quantization error and the resulting drop in accuracy.
These significant enhancements come at almost no cost and are parameter-free, i.e. only a fixed random measurement matrix and a fixed random pre-quantization bias have been introduced while the trainable scaling factor has been eliminated.
As a positive side-effect, the QCS paradigm allows us to quantize the DNN weights and, at the same time, reduce their dimensionality.
In an extended version of this work, numerical tests on various benchmark tasks and datasets will be conducted to validate the above-mentioned benefits.
This will allow the field of deep learning to advance in terms of inference acceleration on resource-limited devices, which will pave the way for a wider and more inclusive adoption of DNN based technologies.

\small


\end{document}